\documentclass[sigplan,screen,nonacm,10pt]{acmart}
\renewcommand\footnotetextcopyrightpermission[1]{}

\setcopyright{none}
\usepackage{amsmath,amsfonts,bm}

\def\eqref#1{equation~\ref{#1}}
\def\1{\bm{1}}

\DeclareMathAlphabet{\mathsfit}{\encodingdefault}{\sfdefault}{m}{sl}
\SetMathAlphabet{\mathsfit}{bold}{\encodingdefault}{\sfdefault}{bx}{n}

\title{\proj: Composable Benchmark Generation to Reduce Deep Learning Benchmarking Effort on CPUs (Extended)}

\PassOptionsToPackage{numbers, compress}{natbib}

\usepackage{booktabs} % For formal tables
\usepackage{lineno, blindtext, color}

\usepackage{tcolorbox}
\usepackage{environ}
\usepackage{xargs}                      % Use more than one optional parameter in a new commands

\usepackage{listings}

\usepackage[htt]{hyphenat}
\usepackage{latexsym}
\usepackage{verbatim}
\usepackage{graphicx}
\usepackage{subcaption} 
\usepackage[font={small}]{caption}
\usepackage{booktabs} % for professional tables
\usepackage{soul}
\usepackage{amsthm}
\usepackage{amsmath}
\usepackage[utf8]{inputenc}
\usepackage{xspace}
\usepackage{array}
\usepackage{makecell}
\usepackage{rotating}
\usepackage{filecontents}
\usepackage{pgfplots}
\usepackage{pgfplotstable}
\usepackage{mathrsfs}
\usepackage[normalem]{ulem}
\usepackage{enumitem}
\usepackage{amssymb}% http://ctan.org/pkg/amssymb

\usepackage{wrapfig}
\usepackage{caption}
\usepackage{subcaption}
\usepackage{times}
\usepackage{url}
\makeatletter
\newsavebox{\measure@tikzpicture}
\NewEnviron{scaletikzpicturetowidth}[1]{%
  \def\tikz@width{#1}%
  \begin{lrbox}{\measure@tikzpicture}%
  \BODY
  \end{lrbox}%
  \pgfmathparse{#1/\wd\measure@tikzpicture}%
  \BODY
}

\setlist{noitemsep,nolistsep}

\newcommand{\cmmnt}[1]{\ignorespaces}

\definecolor{myred}{rgb}{0.843137,0.188235,0.152941}
\definecolor{myblack}{rgb}{0.27451,0.32549,0.384314}
\definecolor{mygreen}{rgb}{0.301961,0.686275,0.290196}
\definecolor{myyellow}{rgb}{0.996078,0.878431,0.564706}
\definecolor{myblue}{rgb}{0.568627,0.74902,0.858824}

\pgfplotsset{compat=newest,}
\usepgfplotslibrary{external}
\usetikzlibrary{babel}
\pgfplotsset{every axis/.style={scale only axis}}

\definecolor{plotcolor1}{rgb}{0.568627,0.74902,0.858824}
\definecolor{plotcolor2}{rgb}{0.996078,0.878431,0.564706}
\definecolor{plotcolor3}{rgb}{0.27451,0.32549,0.384314}
\definecolor{plotcolor4}{rgb}{0.843137,0.188235,0.152941}
\definecolor{plotcolor5}{rgb}{0.988235,0.552941,0.34902}
\definecolor{plotcolor6}{rgb}{0.596078,0.305882,0.639216}
\definecolor{plotcolor7}{rgb}{0.65098,0.337255,0.156863}
\definecolor{plotcolor8}{rgb}{0.105882,0.619608,0.466667}
\definecolor{plotcolor9}{rgb}{1.,1.,0.6}
\definecolor{plotcolor10}{rgb}{0.745098,0.729412,0.854902}
\definecolor{plotcolor11}{rgb}{0.984314,0.501961,0.447059}
\definecolor{plotcolor12}{rgb}{0.501961,0.694118,0.827451}
\definecolor{plotcolor13}{rgb}{1.,1.,0.2}
\definecolor{plotcolor14}{rgb}{0.992157,0.705882,0.384314}
\definecolor{plotcolor15}{rgb}{0.988235,0.803922,0.898039}
\definecolor{plotcolor16}{rgb}{0.701961,0.870588,0.411765}
\definecolor{plotcolor17}{rgb}{0.215686,0.494118,0.721569}
\definecolor{plotcolor18}{rgb}{0.941176,0.231373,0.12549}
\definecolor{plotcolor19}{rgb}{0.168627,0.54902,0.745098}
\definecolor{plotcolor20}{rgb}{0.552941,0.827451,0.780392}
\definecolor{plotcolor21}{rgb}{0.968627,0.505882,0.74902}
\definecolor{plotcolor22}{rgb}{0.6,0.6,0.6}
\definecolor{plotcolor23}{rgb}{0.301961,0.686275,0.290196}
\definecolor{plotcolor24}{rgb}{0.980392,0.501961,0.447059}

\usepgfplotslibrary{colorbrewer}
\pgfplotsset{cycle list/Dark2-8}
\newcommand{\ignore}[1]{}

\DeclareRobustCommand*\circled[1]{\tikz[baseline=(char.base)]{
            \node[shape=circle,white,draw,fill=black,inner sep=0.0pt] (char) {\small #1};}}
\newcommand*\circledwhite[1]{\tikz[baseline=(char.base)]{
            \node[shape=circle,draw,inner sep=0.0pt] (char) {\small  #1};}}
\newcolumntype{C}{>{\centering\arraybackslash} m{.1\linewidth} }  %# New column type

\usepackage{times}
\usepackage{epsfig}
\usepackage{graphicx}
\usepackage{amsmath}
\usepackage{amssymb}
\usepackage{enumitem}
\usepackage{enumitem}
\usepackage{kantlipsum}
\usepackage{adjustbox}

\usepackage[utf8]{inputenc} % allow utf-8 input
\usepackage[T1]{fontenc}    % use 8-bit T1 fonts
\usepackage{hyperref}       % hyperlinks
\usepackage{url}            % simple URL typesetting
\usepackage{booktabs}       % professional-quality tables
\usepackage{amsfonts}       % blackboard math symbols
\usepackage{nicefrac}       % compact symbols for 1/2, etc.
\usepackage{microtype}      % microtypography
\usepackage{arydshln}
\usepackage{xspace}
\usepackage{filecontents}
\usepackage{pgfplots}
\usepackage{pgfplotstable}
\usepackage{multirow}
\usepackage{tikz}
\usepackage[normalem]{ulem}
\usepackage{enumitem}
\usepackage{listings}
\usepackage{tcolorbox}
\usepackage{xargs}                      % Use more than one optional
\usepackage{amssymb}% http://ctan.org/pkg/amssymb
\usepackage{pifont}% http://ctan.org/pkg/pifont

\usepackage{environ}% http://ctan.org/pkg/environ

\tcbuselibrary{listings,breakable}

\definecolor{myred}{rgb}{0.843137,0.188235,0.152941}
\definecolor{myblack}{rgb}{0.27451,0.32549,0.384314}
\definecolor{mygreen}{rgb}{0.301961,0.686275,0.290196}
\definecolor{myyellow}{rgb}{0.996078,0.878431,0.564706}
\definecolor{myblue}{rgb}{0.568627,0.74902,0.858824}

\input{glyphtounicode}
\DeclareRobustCommand*\circled[1]{\tikz[baseline=(char.base)]{
            \node[circle,white,draw,fill=black,inner sep=1.0pt] (char) {\normalfont\footnotesize\sffamily\textsf{#1}};}}
\DeclareRobustCommand*\circledwhite[1]{\tikz[baseline=(char.base)]{
            \node[circle,line width=0.5mm,draw,inner sep=1.0pt] (char) {\normalfont\footnotesize\sffamily\textsf{#1}};}}

\definecolor{circlered}{RGB}{176, 0, 29}

\DeclareRobustCommand*\circledwhitered[1]{\tikz[baseline=(char.base)]{
            \node[shape=circle,line width=0.5mm,circlered,draw,text=black,inner sep=0.5pt,anchor=base] (char) {\normalfont\footnotesize\sffamily\textsf{#1}};}}

\usepackage{algorithmicx}
\usepackage{algorithm}
\usepackage{algpseudocode}

\algnewcommand{\IIf}[1]{\State\algorithmicif\ #1\ \algorithmicthen}
\algnewcommand{\EndIIf}{\unskip\ \algorithmicend\ \algorithmicif}

\algblock[TryCatchFinally]{try}{endtry}
\algcblock[TryCatchFinally]{TryCatchFinally}{finally}{endtry}
\algcblockdefx[TryCatchFinally]{TryCatchFinally}{catch}{endtry}
	[1]{\textbf{catch} #1}
	{\textbf{end try}}
	
\algrenewcommand{\Return}{\State\algorithmicreturn~}

\let\OldStatex\Statex
\renewcommand{\Statex}[1][3]{%
  \setlength\@tempdima{\algorithmicindent}%
  \OldStatex\hskip\dimexpr#1\@tempdima\relax}

\usepackage{tcolorbox}

\newtcolorbox{observationbox}[1][]
{
    breakable,
    left=1pt,
    right=1pt,
    top=1pt,
    bottom=1pt,
    colback=gray!20,
    colframe=black,
    width=\dimexpr\columnwidth\relax,
    enlarge left by=0mm,
    boxsep=5pt,
    arc=0pt,outer arc=0pt,
    #1
}

\newtcbox\observationti{hbox, on line, colback=black, enhanced, frame hidden, boxrule=0pt, 
    top=-2pt, bottom=-2pt, right=-2pt, left=-2pt, rounded corners, arc=2pt}

\newcommand{\proj}{DLBricks\xspace}

\begin{document}

\author{Cheng Li}
\affiliation{%
  \institution{University of Illinois Urbana-Champaign}
  \city{Urbana}
  \state{Illinois}
}
\email{cli99@illinois.edu}

\author{Abdul Dakkak}
\affiliation{%
  \institution{University of Illinois Urbana-Champaign}
  \city{Urbana}
  \state{Illinois}
}
\email{dakkak@illinois.edu}

\author{Jinjun Xiong}
\affiliation{%
  \institution{IBM T. J. Watson Research Center}
  \city{Yorktown Heights}
  \state{New York}
}
\email{jinjun@us.ibm.com}

\author{Wen-mei Hwu}
\affiliation{%
  \institution{University of Illinois Urbana-Champaign}
  \city{Urbana}
  \state{Illinois}
}
\email{w-hwu@illinois.edu}

% % The default list of authors is too long for headers}
\renewcommand{\shortauthors}{Cheng Li, et al.}

% \settopmatter{printfolios=true}

% \thispagestyle{plain}
% \pagestyle{plain}

\begin{abstract}

The past few years have seen a surge of applying Deep Learning (DL) models for a wide array of tasks such as
image classification, object detection, machine translation, etc.
While DL models provide an opportunity to solve otherwise intractable tasks, their adoption relies on them being optimized to meet latency and resource requirements.
Benchmarking is a key step in this process but has been hampered in part due to the lack of representative and up-to-date benchmarking suites.
This is exacerbated by the fast-evolving pace of DL models.

This paper proposes \proj{}, a composable benchmark generation design that reduces the effort of developing, maintaining, and running DL benchmarks on CPUs.
\proj{} decomposes DL models into a set of unique runnable networks and constructs the original model's performance using the performance of the generated benchmarks.
\proj{} leverages two key observations: DL layers are the performance building blocks of DL models and layers are extensively repeated within and across DL models.
Since benchmarks are generated automatically and the benchmarking time is minimized,
\proj can keep up-to-date with the latest proposed models, relieving the pressure of selecting representative DL models.
Moreover, \proj allows users to represent proprietary models within benchmark suites.
We evaluate \proj{} using $50$ MXNet models spanning $5$ DL tasks on $4$ representative CPU systems.
We show that \proj{} provides an accurate performance estimate for the DL models and reduces the benchmarking time across systems (e.g. within $95\%$ accuracy and up to $4.4\times$ benchmarking time speedup on Amazon EC2 \texttt{c5.xlarge}).

\end{abstract}

\maketitle

\section{Introduction}\label{sec:intro}

The recent impressive progress made by Deep Learning (DL) in a wide array of applications, such as autonomous vehicles, face recognition, object detection, machine translation, fraud detection, etc. has led to increased public usage of DL models.
Benchmarking these trained DL models before deployment is critical for DL models to be usable and meet latency and resource constraints. %
Hence there have been significant efforts to develop benchmark suites that evaluate widely used DL models~\cite{mlperf,dawnbench,adolf2016fathom,tbd,aimatrix}.
An example is MLPerf~\cite{mlperf}, which is formed as a collaboration between industry and academia and aims to provide reference implementations for DL model training and inference.
However, the current benchmarking practice has a few limitations that are exacerbated by the fast-evolving pace of DL models.

\begin{figure}
	\centering
	\includegraphics[width=0.48\textwidth]{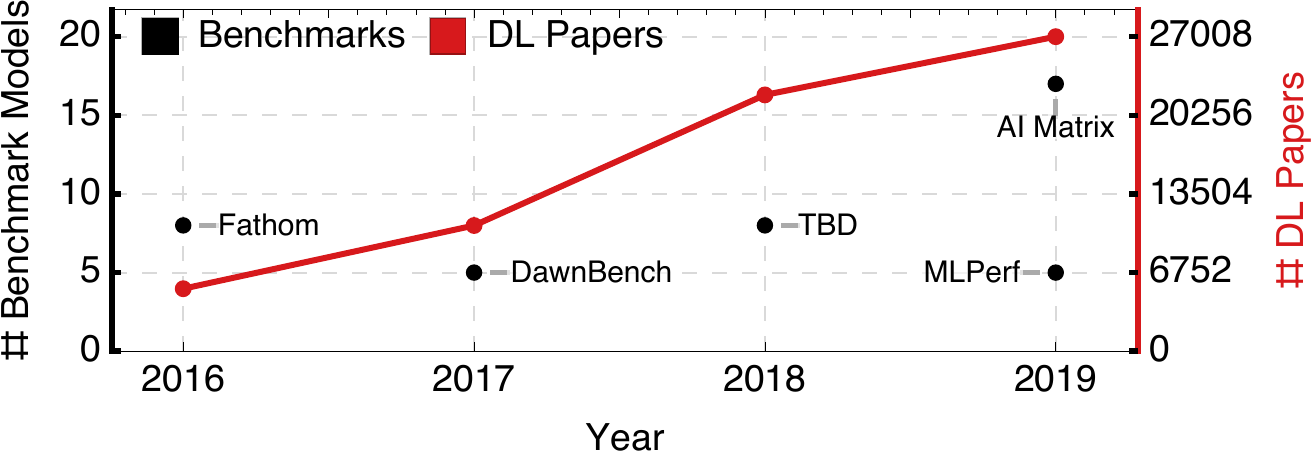}
	\caption{The number of DL models included in the the recent published DL benchmark suites (Fathom~\cite{adolf2016fathom}, DawnBench~\cite{dawnbench}, TBD~\cite{tbd}, AI Matrix~\cite{aimatrix}, and MLPerf~\cite{mlperf}) compared to the number of DL papers published in the same year (using Scopus Preview~\cite{scopus}) .}
	\label{fig:num_benchmarks}
	\vspace{-5pt}
\end{figure}

Developing, maintaining, and running benchmarks takes a non-trivial amount of effort.
Thus, even though there are many benchmark suites, the number of DL model benchmarks within them is small.
For each DL task of interest, benchmark suite authors select a small representative subset (or one) out of tens or even hundreds of candidate models.
Deciding on a representative set of models is an arduous effort as it takes a long debating process to determine what models to add and what to exclude.
For example, it took over a year of weekly discussion to determine and publish MLPerf v$0.5$ inference models, and the number of models was reduced from the $10$  models originally considered to $5$.
Figure~\ref{fig:num_benchmarks} shows the gap between the number of DL papers~\cite{scopus} and the number of models included in recent benchmarking efforts.
Given that DL models are proposed or updated on a daily basis~\cite{dean2018new, hazelwood2018applied}, it is very challenging for benchmark suites to be agile and representative of real-world DL model use.

Moreover, only public available models are considered for inclusion in benchmark suites.
Proprietary models are trade secret or restricted by copyright and cannot be shared externally for benchmarking.
Thus, proprietary models are not included or represented within benchmark suites.
From the proprietary model owners' point of view, since they cannot directly share their models, to benchmark the model using a vendor's system means investing time and money in purchasing vendor hardware, setting it up, and evaluating the model on the newly purchased system.
This is a cumbersome process and places a large burden on proprietary model owners evaluating  SW/HW stacks.
Moreover, the research community cannot collaborate to propose potential optimizations for these proprietary models.

To address the above issues, we propose \proj{} --- a composable benchmark generation design that reduces the effort to develop, maintain, and run DL benchmarks on CPUs.
We focus on latency-sensitive (batch size=$1$) DL inference using CPUs since CPUs are a common and cost-effective option for DL model deployment.
Given a set of DL models, \proj{} parses them into a set of atomic unique layer sequences based on the user-specified benchmark granularity ($G$).
A \textit{layer sequence} is a chain of layers and two layer sequences are considered \textit{non-unique} if they are identical ignoring their weight values.
\proj{} then generates unique (non-overlapping) \textit{runnable networks} (or subgraphs of the model with at most $G$ layers that can be executed by a framework) using the layer sequences' information, and these networks form the representative set of benchmarks for the input models.
Users can run the generated benchmarks on a system of interest and \proj{} uses the benchmark results to construct an estimate of the input models' performance on that system.

\proj{} leverages two key observations on DL inference: \circled{1} Layers are the performance building blocks of the model performance and a simple summation is effective given the current DL software stack (no parallel execution of data-independent layers or overlapping of layer execution) on CPUs.
\circled{2} Layers (considering their layer type, shape, and parameters, but ignoring the weights) are extensively repeated within and across DL models.
\proj{} uses both observations to generate a representative benchmark suite, minimize the time to benchmark, and estimate a model's performance from layer sequences.

Since benchmarks are generated automatically by \proj{}, benchmark development and maintenance effort are greatly reduced.
\proj{} is defined by a set of simple consistent principles and can be used to benchmark and characterize a broad range of models. %
Moreover, since each generated benchmark represents only a subset of the input model, the input model's topology does not appear in the output benchmark suite.
This, along with the fact that ``fake'' or dummy models can be inserted into the set of input models, means that the generated benchmark suite can represent proprietary models without the concern of revealing proprietary models.

In summary, this paper makes the following contributions:
\begin{itemize}[nosep,leftmargin=0.5em,labelwidth=*,align=left]
	\item We perform a comprehensive performance analysis of $50$ state-of-the-art DL models on CPUs and observe that layers are the performance building blocks of DL models, thus a model's performance can be estimated by the performance of its layers.
	\item We also perform an in-depth DL architecture analysis of the DL models and make the observation that DL layers with the same type, shape, and parameters are repeated extensively within and across models.
	\item We propose \proj{}, a composable benchmark generation design on CPUs that decomposes DL models into a set of unique runnable networks and constructs the original model's performance using the performance of the generated benchmarks.
	      \proj{} reduces the effort of developing, maintaining, and running DL benchmarks, and relieves the pressure of selecting representative DL models.
	\item \proj{} allows representing proprietary models without model privacy concerns as the input model's topology does not appear in the output benchmark suite, and ``fake'' or dummy models can be inserted into the set of input models.
	\item We evaluate \proj{} using $50$ MXNet models spanning $5$ DL tasks on $4$ representative CPU systems.
	      We show that \proj{} provides a tight performance estimate for DL models and reduces the benchmarking time across systems.
	      E.g. The composed model latency is within $95\%$ of the model actual performance while up to $4.4\times$ benchmarking speedup is achieved on the Amazon EC2 \texttt{c5.xlarge} system.
\end{itemize}

This paper is structured as follows.
First, in Section~\ref{sec:related} we describe different benchmarking approaches previously performed.
We then (Section~\ref{sec:motiv}) detail two key observations \circled{1} and \circled{2} that enable our design.
We then propose \proj{} in Section~\ref{sec:design} and describe how it provides a streamlined benchmark generation workflow which lowers the time to benchmark.
Section~\ref{sec:eval} evaluates \proj using $50$ models running on $4$ systems.
We then describe future work in Section~\ref{sec:future} before we conclude in Section~\ref{sec:conc}.

\section{Motivation}\label{sec:motiv}

\proj{} is designed based on two key observations presented in this section.
To demonstrate and support these observations, we perform comprehensive performance and architecture analysis of state-of-the-art DL model.
Evaluations in this section use $50$ MXNet models of different DL tasks (listed in Table~\ref{tab:models_full}) and were run with MXNet (v$1.5.1$ MKL release) on a Amazon \texttt{c5.2xlarge} instance (as listed in Table~\ref{tab:systems}).
We focus on latency sensitive (batch size $=1$) DL inference on CPUs.

\begin{table*}
	\centering
	\resizebox{0.8\textwidth}{!}{%
		\centering
		\begin{tabular}{rllrr} \toprule
			\centering%
			\textbf{\thead{ID}} & \textbf{\thead{Name}}                                                                & \textbf{\thead{Task}} & \textbf{\thead{Num Layers}} \\ \midrule
			$1$                 & Ademxapp Model A Trained on ImageNet Competition Data                                & Classification        & $142$                       \\  %
			$2$                 & Age Estimation VGG-16 Trained on IMDB-WIKI and Looking at People Data                & Classification        & $40$                        \\  %
			$3$                 & Age Estimation VGG-16 Trained on IMDB-WIKI Data                                      & Classification        & $40$                        \\  %
			$4$                 & CapsNet Trained on MNIST Data                                                        & Classification        & $53$                        \\  %
			$5$                 & Gender Prediction VGG-16 Trained on IMDB-WIKI Data                                   & Classification        & $40$                        \\  %
			$6$                 & Inception V1 Trained on Extended Salient Object Subitizing Data                      & Classification        & $147$                       \\  %
			$7$                 & Inception V1 Trained on ImageNet Competition Data                                    & Classification        & $147$                       \\  %
			$8$                 & Inception V1 Trained on Places365 Data                                               & Classification        & $147$                       \\  %
			$9$                 & Inception V3 Trained on ImageNet Competition Data                                    & Classification        & $311$                       \\  %
			$10$                & MobileNet V2 Trained on ImageNet Competition Data                                    & Classification        & $153$                       \\  %
			$11$                & ResNet-101 Trained on ImageNet Competition Data                                      & Classification        & $347$                       \\  %
			$12$                & ResNet-101 Trained on YFCC100m Geotagged Data                                        & Classification        & $344$                       \\
			$13$                & ResNet-152 Trained on ImageNet Competition Data                                      & Classification        & $517$                       \\  %
			$14$                & ResNet-50 Trained on ImageNet Competition Data                                       & Classification        & $177$                       \\  %
			$15$                & Squeeze-and-Excitation Net Trained on ImageNet Competition Data                      & Classification        & $874$                       \\  %
			$16$                & SqueezeNet V1.1 Trained on ImageNet Competition Data                                 & Classification        & $69$                        \\  %
			$17$                & VGG-16 Trained on ImageNet Competition Data                                          & Classification        & $40$                        \\  %
			$18$                & VGG-19 Trained on ImageNet Competition Data                                          & Classification        & $46$                        \\  %
			$19$                & Wide ResNet-50-2 Trained on ImageNet Competition Data                                & Classification        & $176$                       \\  %
			$20$                & Wolfram ImageIdentify Net V1                                                         & Classification        & $232$                       \\  %
			$21$                & Yahoo Open NSFW Model V1                                                             & Classification        & $177$                       \\  \hdashline
			$22$                & AdaIN-Style Trained on MS-COCO and Painter by Numbers Data                           & Image Processing      & $109$                       \\  %
			$23$                & Colorful Image Colorization Trained on ImageNet Competition Data                     & Image Processing      & $58$                        \\  %
			$24$                & ColorNet Image Colorization Trained on ImageNet Competition Data                     & Image Processing      & $62$                        \\  %
			$25$                & ColorNet Image Colorization Trained on Places Data                                   & Image Processing      & $62$                        \\  %
			$26$                & CycleGAN Apple-to-Orange Translation Trained on ImageNet Competition Data            & Image Processing      & $94$                        \\  %
			$27$                & CycleGAN Horse-to-Zebra Translation Trained on ImageNet Competition Data             & Image Processing      & $94$                        \\  %
			$28$                & CycleGAN Monet-to-Photo Translation                                                  & Image Processing      & $94$                        \\  %
			$29$                & CycleGAN Orange-to-Apple Translation Trained on ImageNet Competition Data            & Image Processing      & $94$                        \\  %
			$30$                & CycleGAN Photo-to-Cezanne Translation                                                & Image Processing      & $96$                        \\  %
			$31$                & CycleGAN Photo-to-Monet Translation                                                  & Image Processing      & $94$                        \\  %
			$32$                & CycleGAN Photo-to-Van Gogh Translation                                               & Image Processing      & $96$                        \\  %
			$33$                & CycleGAN Summer-to-Winter Translation                                                & Image Processing      & $94$                        \\  %
			$34$                & CycleGAN Winter-to-Summer Translation                                                & Image Processing      & $94$                        \\  %
			$35$                & CycleGAN Zebra-to-Horse Translation Trained on ImageNet Competition Data             & Image Processing      & $94$                        \\  %
			$36$                & Pix2pix Photo-to-Street-Map Translation                                              & Image Processing      & $56$                        \\  %
			$37$                & Pix2pix Street-Map-to-Photo Translation                                              & Image Processing      & $56$                        \\  %
			$38$                & Very Deep Net for Super-Resolution                                                   & Image Processing      & $40$                        \\ \hdashline %
			$39$                & SSD-VGG-300 Trained on PASCAL VOC Data                                               & Object Detection      & $145$                       \\  %
			$40$                & SSD-VGG-512 Trained on MS-COCO Data                                                  & Object Detection      & $157$                       \\  %
			$41$                & YOLO V2 Trained on MS-COCO Data                                                      & Object Detection      & $106$                       \\ \hdashline %
			$42$                & 2D Face Alignment Net Trained on 300W Large Pose Data                                & Regression            & $967$                       \\  %
			$43$                & 3D Face Alignment Net Trained on 300W Large Pose Data                                & Regression            & $967$                       \\  %
			$44$                & Single-Image Depth Perception Net Trained on Depth in the Wild Data                  & Regression            & $501$                       \\  %
			$45$                & Single-Image Depth Perception Net Trained on NYU Depth V2 and Depth in the Wild Data & Regression            & $501$                       \\  %
			$46$                & Single-Image Depth Perception Net Trained on NYU Depth V2 Data                       & Regression            & $501$                       \\  %
			$47$                & Unguided Volumetric Regression Net for 3D Face Reconstruction                        & Regression            & $1029$                      \\ \hdashline %
			$48$                & Ademxapp Model A1 Trained on ADE20K Data                                             & Semantic Segmentation & $141$                       \\  %
			$49$                & Ademxapp Model A1 Trained on PASCAL VOC2012 and MS-COCO Data                         & Semantic Segmentation & $141$                       \\  %
			$50$                & Multi-scale Context Aggregation Net Trained on CamVid Data                           & Semantic Segmentation & $53$                        \\  %
			\bottomrule
		\end{tabular}%
	}
	\caption{We use $50$ MXNet models from the publicly available Wolfram Neural Net Repository~\cite{wolfram} for evaluation.
		The models span across $5$ DL tasks.
		All models are run using MXNet.
		The number of layers in a model is listed.}
	\label{tab:models_full}
	\vspace{-10pt}
\end{table*}

\subsection{Layers as the Performance Building Blocks}\label{sec:perf}

A DL model is a directed acyclic graph (DAG) where each vertex within the DAG is a layer (or operator, such as convolution, batchnormalization, pooling, element-wise, softmax) and an edge represents the transfer of data.
For a DL model, a \textit{layer sequence} is defined as a simple path within the DAG containing one or more vertices.
A \textit{subgraph}, on the other hand,  is defined as a DAG composed by one or more layers within the model (i.e. subgraph is a superset of layer sequence, and may or may not be a simple path).
We are only interested in network subgraphs that are runnable within frameworks and we call these runnable subgraphs \textit{runnable networks}.

\begin{figure*}[t]
	\centering
	\begin{subfigure}[b]{\textwidth}
		\centering
		\includegraphics[width=\textwidth]{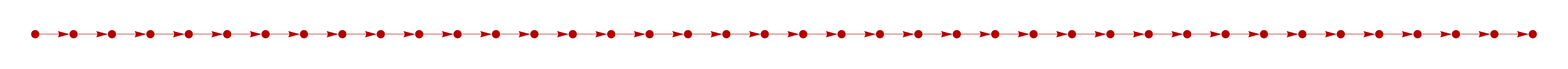}
		\caption{
			\texttt{VGG16} (ID=$17$).
		}
		\label{fig:vgg16_critical}
	\end{subfigure}\\
	\begin{subfigure}[b]{\textwidth}
		\centering
		\setlength{\abovecaptionskip}{-10pt}
		\includegraphics[width=\textwidth]{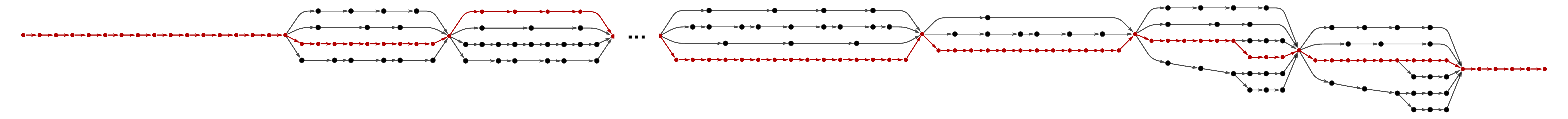}
		\caption{
			\texttt{Inception V3} (ID=$9$).
		}
		\label{fig:inception_v3_critical}
	\end{subfigure}%
	\hfill%
	\caption{The model architecture of \texttt{VGG16} (ID=$17$) and \texttt{Inception V3} (ID=$9$). The critical path in each model is highlighted in red. }
	\label{fig:model_arch}
\end{figure*}

DL models may contain layers that can be executed independently in parallel.
The runnable network determined by these data-independent layers is called a \textit{parallel module}.
For example, Figure~\ref{fig:vgg16_critical} shows the \texttt{VGG16}~\cite{DBLP:journals/corr/SimonyanZ14a} (ID=$17$)  the model architecture.
\texttt{VGG16} contains no parallel modules and is a linear sequence of layers.
\texttt{Inception V3}~\cite{szegedy2016rethinking} (ID=$9$) (shown in Figure ~\ref{fig:inception_v3_critical}), on the other hand, contains  a mix of layer sequences and  parallel-modules.

\begin{figure*}
	\centering
	\includegraphics[width=\textwidth]{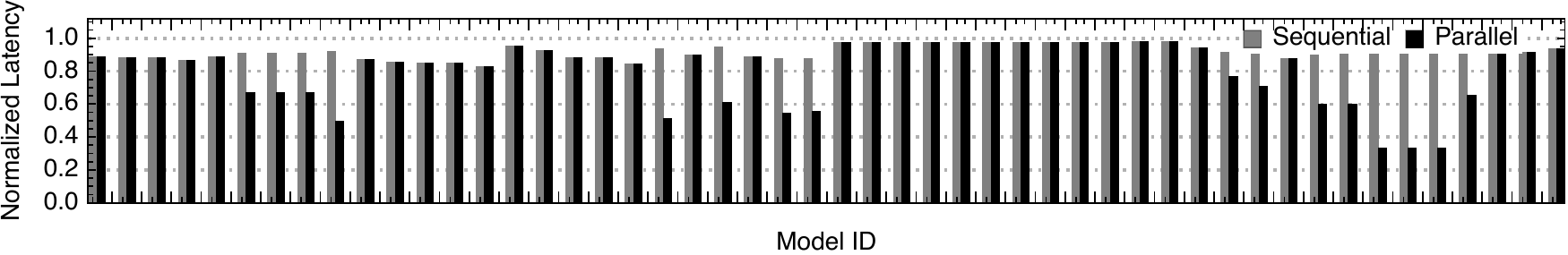}
	\caption{
		The sequential and parallel total layer latency normalized to the model's end-to-end latency using batch size $1$ on \texttt{c5.2xlarge}.
	}
	\label{fig:sequential_vs_parallel}
\end{figure*}

DL frameworks  such as TensorFlow~\cite{tensorflow-paper-2016}, PyTorch~\cite{paszke2017pytorch}, and MXNet~\cite{mxnet} execute a DL model by running the layers within the model graph.
We explore the relation between layer performance and model performance by decomposing each DL model in Table~\ref{tab:models_full} into layers. %
We define a model's \textit{critical path} to be a simple path from the start to the end layer with the highest latency.
For a DL model, we add all its layers' latency and refer to the sum as the \textit{sequential total layer latency}, since this assumes all the layers are executed sequentially by the DL framework.
Theoretically, data-independent paths within the parallel module can be executed in parallel, thus we also calculate
the \textit{parallel total layer latency} by adding up the layer latencies along the critical path.
The critical path of both \texttt{VGG 16} (ID=$17$) and \texttt{Inception V3} (ID=$9$) is highlighted in red in Figure~\ref{fig:model_arch}.
For models that do not have parallel modules, the sequential total layer latency is equal to the parallel total layer latency.

For each of the $50$ models, we compare both sequential and parallel total layer latency to the model's end-to-end latency.
Figure~\ref{fig:sequential_vs_parallel} shows the normalized latencies in both cases.
For models with parallel modules, the parallel total layer latencies are much lower than the model's end-to-end latency.
The difference between the sequential total layer latencies and the models' end-to-end latencies are small.
The normalized latencies are close to $1$ with a geometric metric mean of $91.8\%$ for the sequential case.
This suggests the current software/hardware stack does not exploit parallel execution of data-independent layers or overlapping of layer execution, we verified this by inspecting the source code of popular frameworks such as MXNet, PyTorch, and TensorFlow.

Based on the above analysis, we make the \circled{1} observation:
\begin{observationbox}
	\textbf{Observation 1:}
	DL layers are the performance building blocks of the model performance, therefore, a model's performance can be estimated by the performance of its layers.
	Moreover, a simple summation is effective given the current DL software stack (no parallel execution of data-independent layers or overlapping of layer execution) on CPUs.
\end{observationbox}

\subsection{Layer Repeatability}\label{sec:sharing}

From a model architecture point of view, DL layers are identified by its type, shape, parameters.
For example, a convolution layer is identified by its input shape, output channels, kernel size, stride, padding, dilation, etc.
Layers with the same type, shape, parameters (i.e. only differ in weights) are expected to have the same performance.
We inspected the sources code of popular framework and verified this, as they do not perform any special optimizations for weights.
Thus in this paper we consider two layers to be the \textit{same} if they have the same type, shape, parameters, ignoring weight values, and two layers are \textit{unique} if they are not the same.

\begin{figure*}
	\centering
	\includegraphics[width=\textwidth]{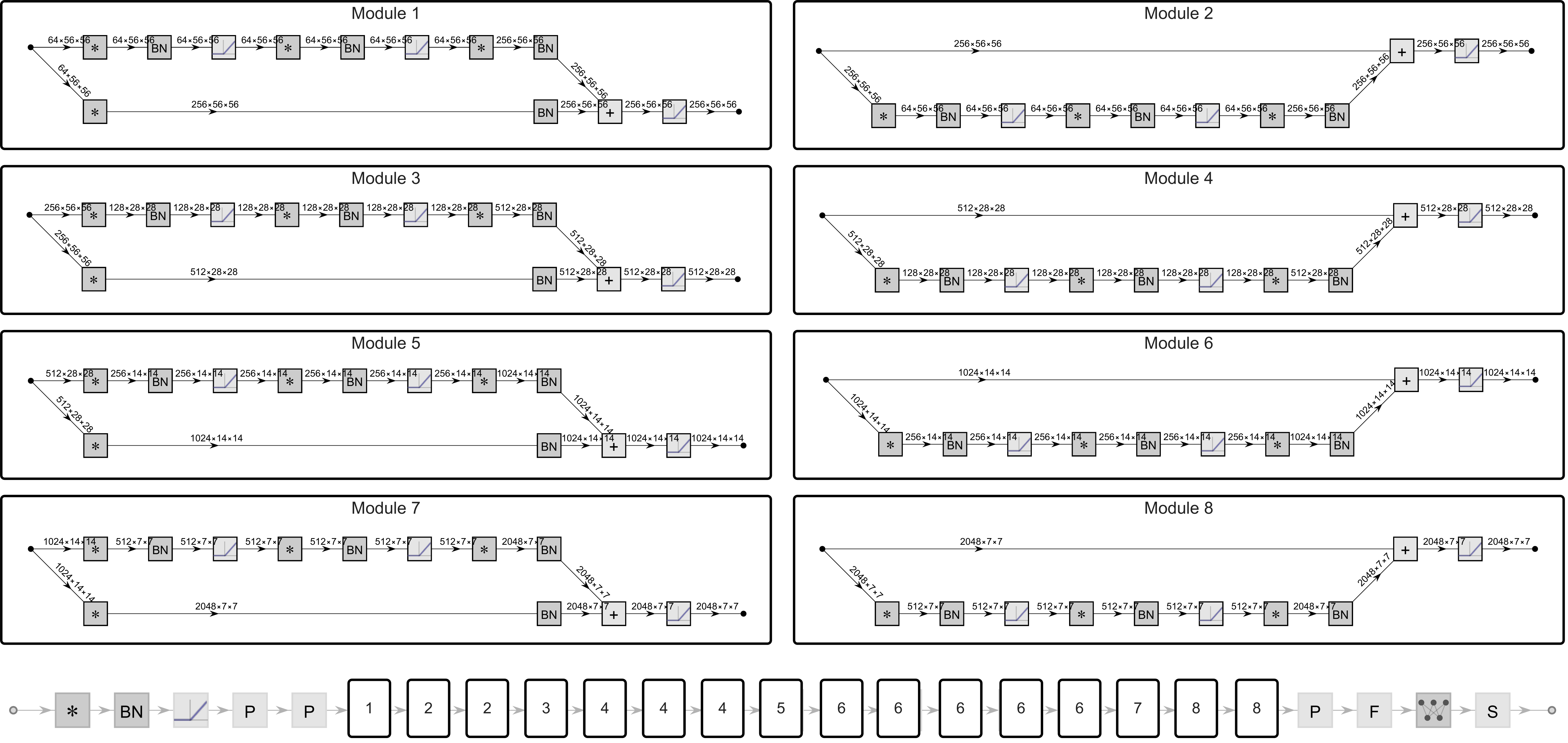}
	\caption{
		The \texttt{ResNet-50} (ID=$14$) architecture. The detailed ResNet modules $1-8$ are listed above the model graph.
	}
	\label{fig:resnet50_v1}
\end{figure*}

DL models tend to have repeated layers or modules (or subgraphs, e.g. Inception and ResNet modules).
For example, Figure~\ref{fig:resnet50_v1} shows the model architecture of \texttt{ResNet-50} with the ResNet module detailed.
Different ResNet modules have layers in common and ResNet module $2, 4, 6, 8$ are entirely repeated within \texttt{ResNet-50}.
Moreover, DL models are often built on top of existing models (e.g. transfer learning~\cite{weiss2016survey} where model are retrained with different data), using common modules (e.g. TensorFlow Hub~\cite{tensorflowhub}), or using layer bundles for Neural Architecture Search~\cite{elsken2019neural,dai2019chamnet,DBLP:journals/corr/abs-1812-03443}.
This results in ample repeated layers when looking at a corpus of models.
We quantitatively explore the layer repeatability within and across models.

\begin{figure*}
	\centering
	\includegraphics[clip,width=\textwidth]{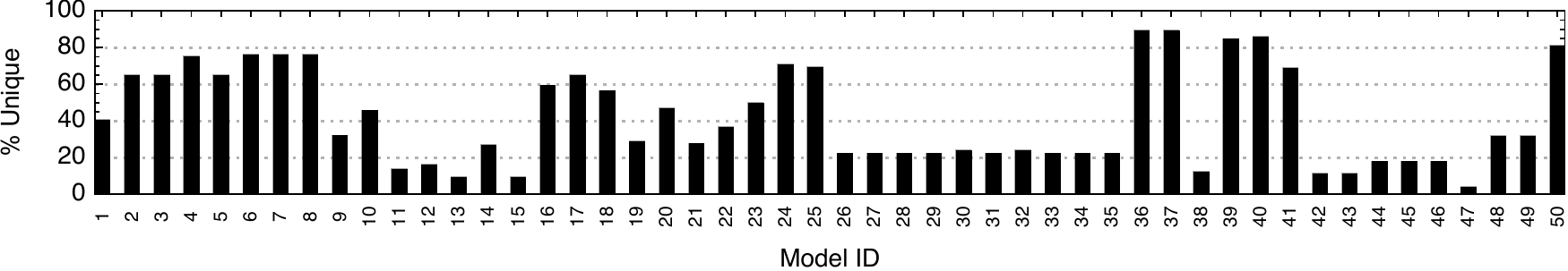}
	\caption{The percentage of unique layers in models in Table~\ref{tab:models_full}, indicating that some layers are repeated within the model.}
	\label{fig:within_model_sharing}
\end{figure*}

\begin{figure*}
	\centering
	\includegraphics[clip,width=\textwidth]{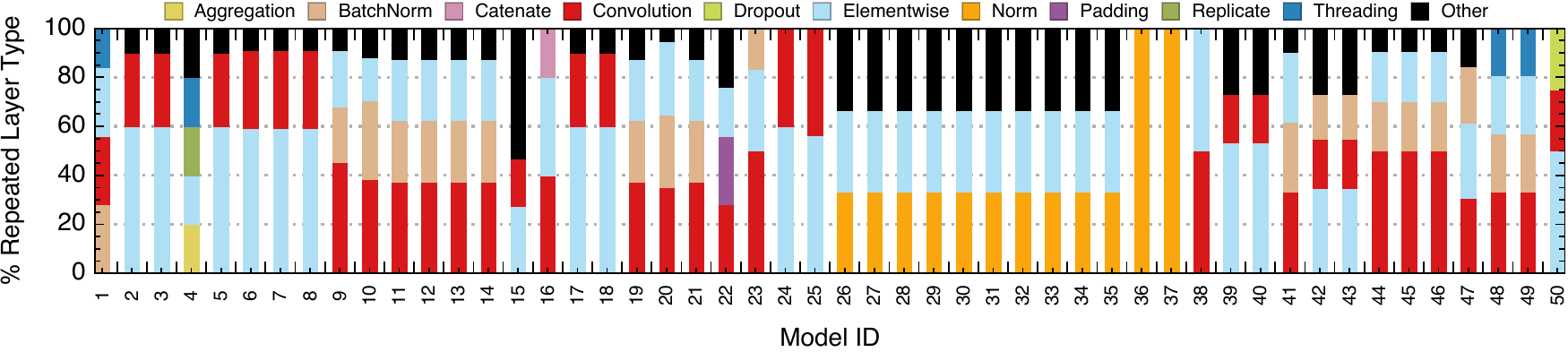}
	\caption{The type distribution of the repeated layers within each model.}
	\label{fig:within_model_no_sharing}
\end{figure*}

Figure~\ref{fig:within_model_sharing} shows the percentage of unique layers within each model in Table~\ref{tab:models_full}.
We can see that layers are extensively repeated within DL models.
For example, in \texttt{Unguided Volumetric Regression Net for 3D Face Reconstruction} (ID=$47$) which has $1029$ layers, only $3.9\%$ of the total layers are unique.
We further look at the repeated layers within each model and Figure~\ref{fig:within_model_no_sharing} shows their type distribution.
As we can see Convolution, Elementwise, BatchNorm, and Norm are the most repeated layer types in terms of intra-model layer repeatability.
If we consider all $50$ models in  Table~\ref{tab:models_full}, the total number of layers is $10815$, but only $1529$ are unique (i.e. $14\%$ are unique).

\begin{figure}
	\centering
	\includegraphics[clip,width=0.48\textwidth]{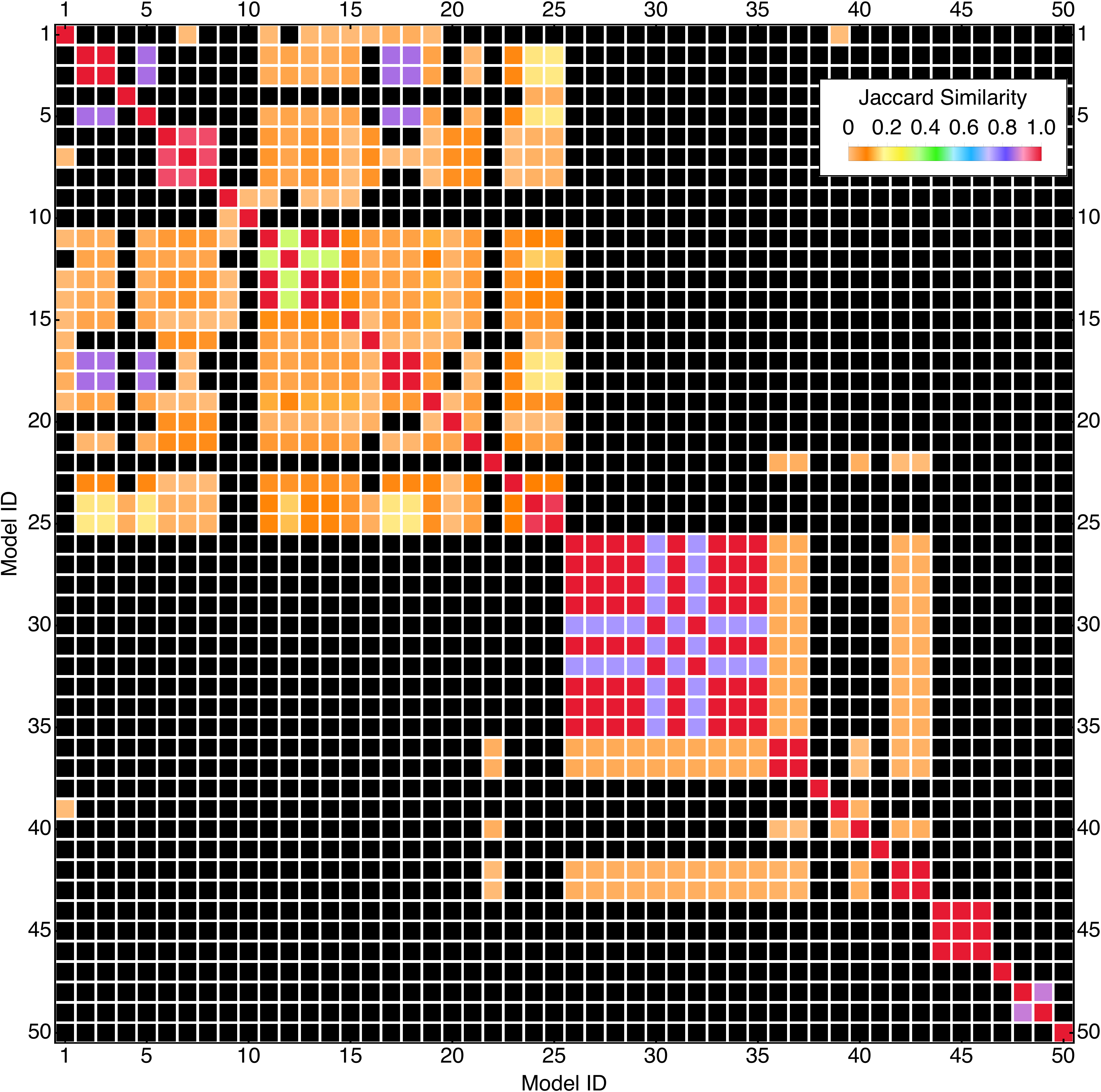}
	\caption{The Jaccard Similarity grid of the models in Table~\ref{tab:models_full}. Each cell corresponds to the Jaccard similarity coefficient between the models at the row and column. Solid red indicates two models have identical layers, and black means there is no common layer.}
	\label{fig:across_model_sharing}
\end{figure}

We illustrate the layer repeatability across models by quantifying the similarity of any two models listed in Table~\ref{tab:models_full}.
We use the Jaccard similarity coefficient; i.e. for any two models $M_1$ and $M_2$ the Jaccard similarity coefficient is defined by $\frac{\left | \mathcal{L}_1 \cap \mathcal{L}_2 \right |}{\left | \mathcal{L}_1 \cup \mathcal{L}_2 \right |}$ where $\mathcal{L}_1$ and $\mathcal{L}_2$ are the layers of $M_1$ and $M_2$ respectively.
The results are shown in Figure~\ref{fig:across_model_sharing}.
As shown, models that shared the same base architecture but are retrained using different data (e.g. \texttt{CycleGAN*} models with IDs $26-35$ and \texttt{Inception V1*} models with IDs $6-8$) have many common layers.
Layers are common across models within the same family (e.g. \texttt{ResNet*}) since they are built from the same set of modules (e.g. \texttt{ResNet-50} is shown in Figure~\ref{fig:resnet50_v1}), or when solving the same task (e.g. image classification task category).
Based on the above analysis, we make yet another \circled{2} observation:
\begin{observationbox}
	\textbf{Observation 2:} Layers are repeated within and across DL models.
	This affords us to decrease the time to benchmark since only a representative set of layers need to be evaluated.
\end{observationbox}

The above two observations suggest that if we can decompose models into layers, and then take the union of them to produce a set of representative runnable networks, then benchmarking the representative runnable networks is sufficient to reconstruct the performance of the input models.
Since we only look at the representative set, the total runtime is less than running all models directly, thus \proj{} can be used to reduce benchmarking time.
Since layer decomposition elides the input model topology, models can be private while their benchmarks can be public.
The next section (Section~\ref{sec:design}) describes how we leverage these two observations to build a benchmark generator while having a workflow where one can reconstruct a model's performance based on the benchmarked layer performance.
We further explore the design space of benchmark granularity and its effect on performance construction accuracy.

\begin{figure}
	\centering
	\includegraphics[width=0.35\textwidth]{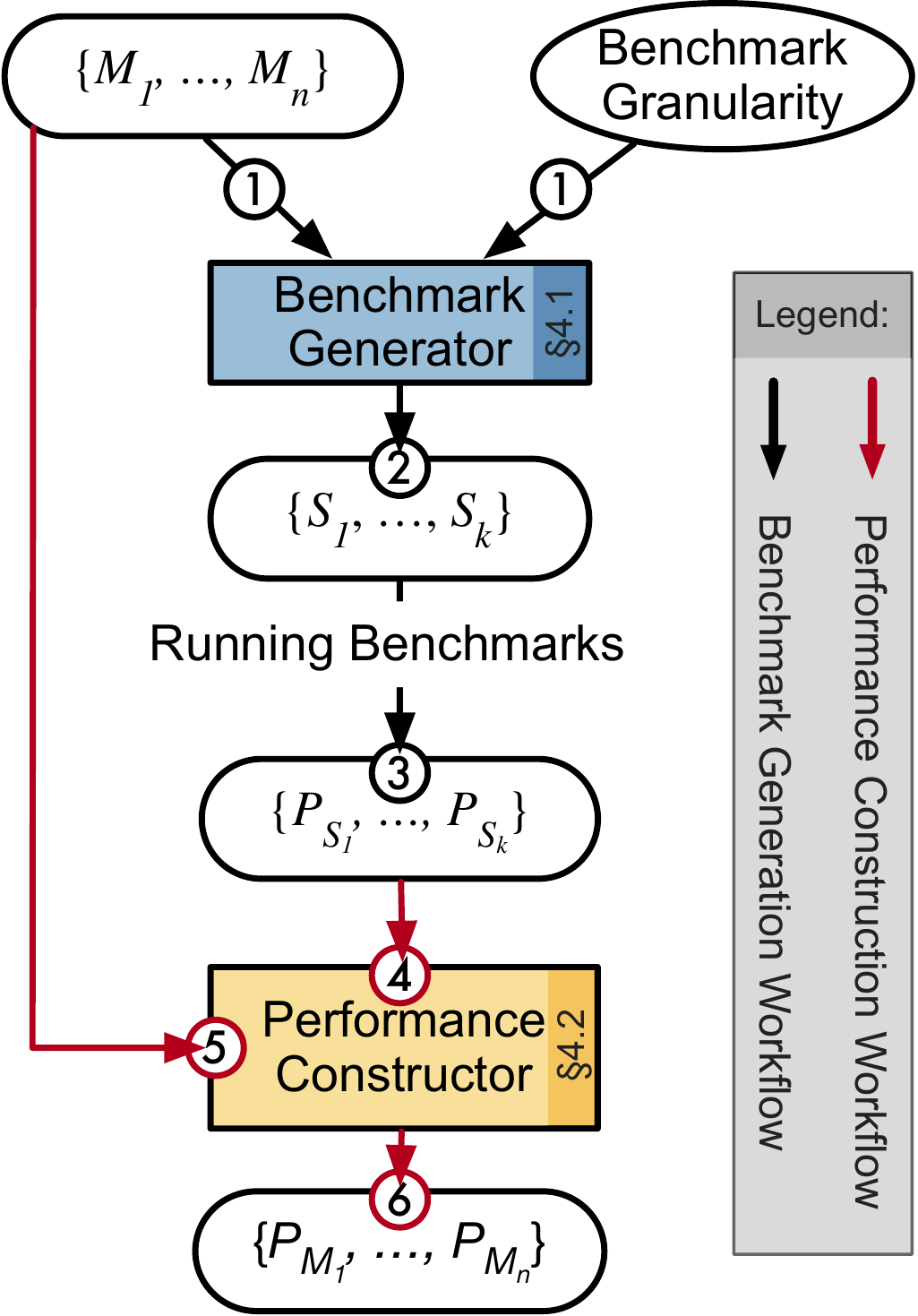}
	\caption{
		\proj{} design and workflow.
	}
	\label{fig:design}
	\vspace{-10pt}
\end{figure}

\section{\proj{} Design}\label{sec:design}

This section presents \proj, a composable benchmark generation design for DL models.
The design is motivated by the two observations discussed in Section~\ref{sec:motiv}.
\proj explores not only layer level model composition but also sequence level composition where a layer sequence is a chain of layers.
The \textit{benchmark granularity} ($G$) specifies the maximum numbers of layers within any layer sequence in the output generated benchmarks.

The design and workflow of \proj is shown in Figure~\ref{fig:design}.
\proj consists of a benchmark generation workflow and a performance construction workflow.
To generate composable benchmarks, one uses the \textit{benchmark generator workflow} where: \circledwhite{1} the user inputs a set of models ($M_1, ..., M_n$) along with a target benchmark granularity.
\circledwhite{2} The benchmark generator parses the input models into a representative (unique) set of non-overlapping layer sequences and then generates a set of runnable networks ($S_1, ..., S_k$) using these layer sequences' information.
\circledwhite{3} The user evaluates the set of runnable networks on a system of interest to get each benchmark's corresponding performance ($P_{S_1}, ..., P_{S_k}$).
The benchmark results are stored and \circledwhitered{4} are used within the \textit{performance construction workflow}.
\circledwhitered{5} To construct the performance of an input model, the performance constructor queries the stored benchmark results for the layer sequences within the model, and then \circledwhitered{6} computes the model's estimated performance ($P_{M_1}, ..., P_{M_k}$).
This section describes both workflows in detail.

\subsection{Benchmark Generator}\label{sec:benchgen}

The benchmark generator takes a list of models $M_1, \ldots, M_n$ and a benchmark granularity $G$.
The \textit{benchmark granularity} specifies the maximum sequence length of the layer sequences generated.
This means that when $G = 1$, each generated benchmark is a single-layer network, whereas when $G=2$ each generated benchmark contains at most $2$ layers.

To split a model with the specified benchmark granularity, we use the \texttt{FindModelSubgraphs} algorithm (Algorithm~\ref{alg:model_subsequence}).
The \texttt{FindModelSubgraphs} takes a model and a maximum sequence length and iteratively generates a set of non-overlapping layer sequences.
First, the layers in the model are sorted topologically and then
calls the \texttt{SplitModel} function (Algorithm~\ref{alg:model_split}) with the desired begin and end layer offset.
This  \texttt{SplitModel} tries to create a runnable DL network using the range of layers desired, if it fails (i.e. a valid DL network cannot be constructed due to input/output layer shape mismatch\footnote{An example invalid network is one which contains a Concat layer, but  does not have all of the Concat layer's required input layers.}), then \texttt{SplitModel} creates a network with the current layer and shifts the begin and end positions by one.
The \texttt{SplitModel} returns a list of created runnable DL networks ($S_i$, \ldots, $S_{i+j}$) along with the end position to \texttt{FindModelSubgraphs}.
The \texttt{FindModelSubgraphs} terminates when no other subsequences can be created.

The benchmark generator applies the  \texttt{FindModelSubgraphs} for each of the input models.
A set of representative (or \textit{unique}) runnable DL networks ($S_1$, \ldots, $S_k$) is then computed.
We say two sequences $S_1$ and $S_2$ are the same if they have the same topology along with the same node parameters (i.e. they are the same DL network modulo the weights).
The unique networks are exported to the frameworks' network format and the user runs them with synthetic input data based on each network's input shape.
The performance of each network is stored ($P_{S_i}$ \ldots, $P_{S_k}$) and used by the performance constructor workflow.

\begin{algorithm}
	\caption{The \texttt{FindModelSubgraphs} algorithm.}
	\label{alg:model_subsequence}
	\begin{flushleft}
		\hspace*{\algorithmicindent} \textbf{Input:} $M$ (Model), $G$ (Benchmark Granularity) \\
		\hspace*{\algorithmicindent} \textbf{Output:}  $Models$
	\end{flushleft}
	\begin{algorithmic}[1]
		\State $begin \gets 0, Models \gets \left \{ \right \}$
		\State $verts \gets \textbf{TopologicalOrder}(\textbf{ToGraph}(M))$
		\While{$begin \le \textbf{Length}(vs)$}
		\State $end \gets  \textbf{Min}(begin + G, \textbf{Length}(vs))$
		\State $sm \gets \texttt{SplitModel}(verts, begin, end)$
		\State $Models \gets Models + sm \left [ \textbf{``models''} \right ]$
		\State $begin \gets sm \left [ \textbf{``end''} \right ] + 1$
		\EndWhile
		\Return $Models$
	\end{algorithmic}
\end{algorithm}

\begin{algorithm}
	\caption{The \texttt{SplitModel} algorithm.}
	\label{alg:model_split}
	\begin{flushleft}
		\hspace*{\algorithmicindent} \textbf{Input:} $verts$, $begin$, $end$\\
		\hspace*{\algorithmicindent} \textbf{Output:} $\left \langle \textbf{``models''}, \textbf{``end''} \right \rangle$ \Comment{Hash table} %
	\end{flushleft}
	\begin{algorithmic}[1]
		\State{$vs \gets verts\left [ begin : end \right ]$}
		\try
		\State{$m \gets \textbf{CreateModel}(vs)$} \Comment{Creates a valid model}
		\Return $\langle \textbf{``models''} \rightarrow \left \{ m \right \}, \textbf{``end''} \rightarrow end \rangle$
		\catch{ModelCreateException}
		\State{$m \gets \left \{ \textbf{CreateModel}(\left \{ verts\left [ begin \right ] \right \}) \right \}$}
		\State{$n \gets \texttt{SplitModel}(verts, begin+1, end+1)$}
		\Return $\langle \textbf{``models''} \rightarrow m+n \left [ \textbf{``models''} \right ],$
		\Statex $\quad\textbf{``end''} \rightarrow n \left [ \textbf{``end''} \right ]  \rangle$
		\endtry
	\end{algorithmic}
\end{algorithm}

\begin{figure*}
	\centering
	\includegraphics[width=\textwidth]{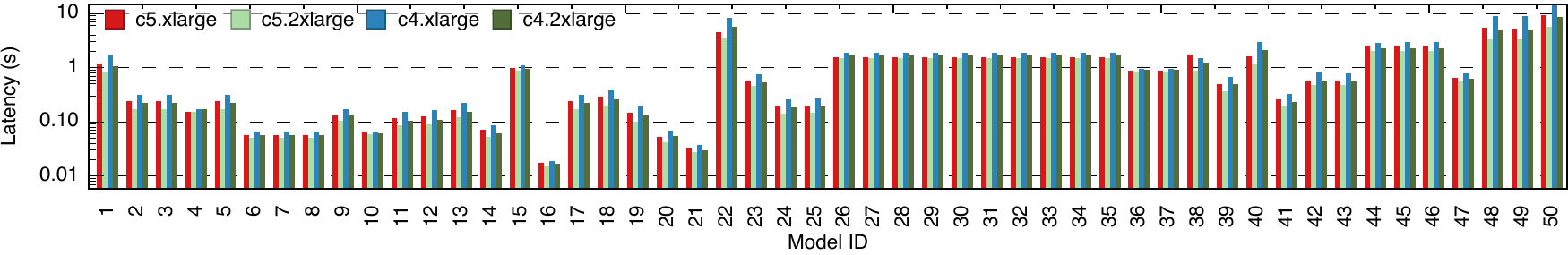}
	\caption{
		The end-to-end latency of all models in log scale across systems.
	}
	\label{fig:end_to_end}

\end{figure*}

\begin{figure*}[ht]
	\centering
	\begin{subfigure}[b]{0.49\textwidth}
		\includegraphics[width=\textwidth]{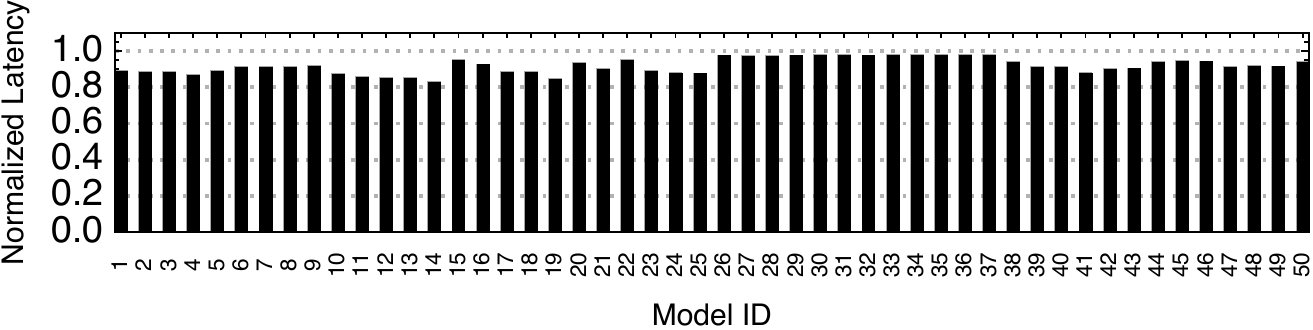}
		\caption{Benchmark Granularity=$1$}
		\label{fig:sequence_length_over_full_time:1}
	\end{subfigure}%
	\hfill
	\begin{subfigure}[b]{0.49\textwidth}
		\includegraphics[width=\textwidth]{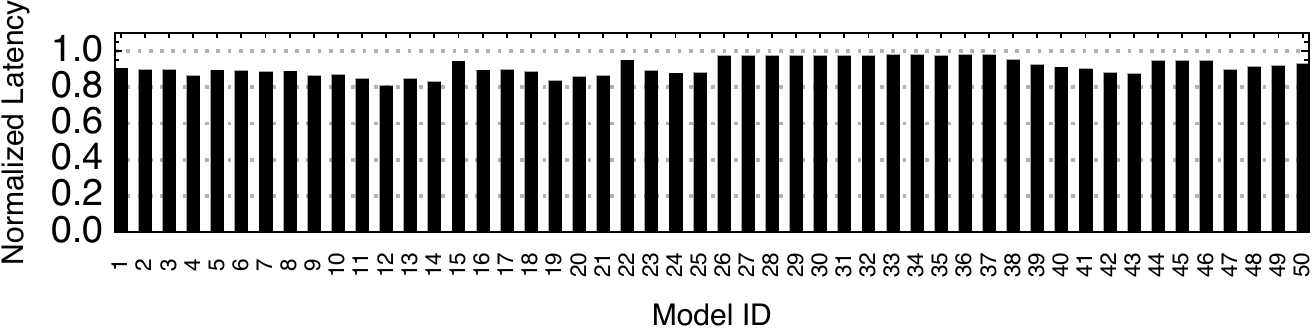}
		\caption{Benchmark Granularity=$2$}
		\label{fig:sequence_length_over_full_time:2}
	\end{subfigure} \\
	\begin{subfigure}[b]{0.49\textwidth}
		\includegraphics[width=\textwidth]{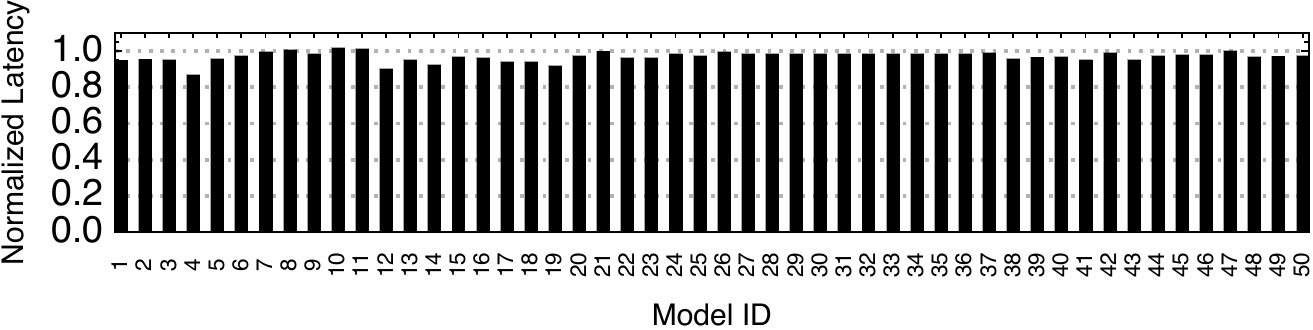}
		\caption{Benchmark Granularity=$3$}
		\label{fig:sequence_length_over_full_time:3}
	\end{subfigure}%
	\hfill
	\begin{subfigure}[b]{0.49\textwidth}
		\includegraphics[width=\textwidth]{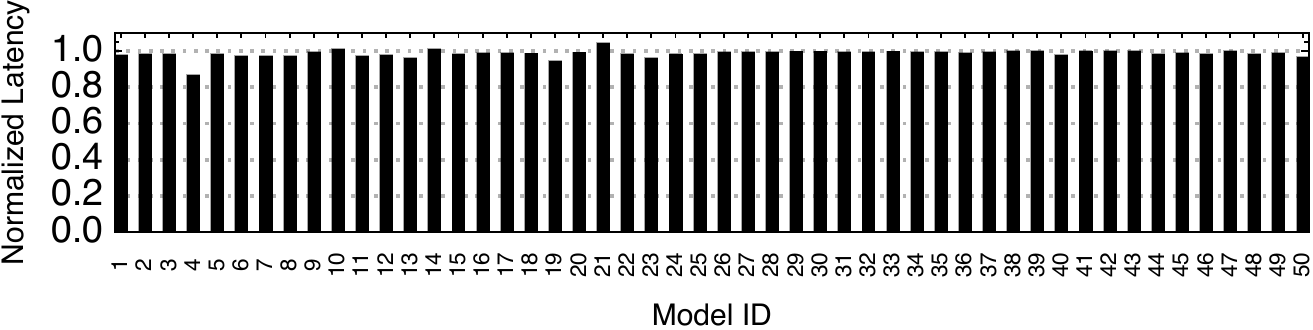}
		\caption{Benchmark Granularity=$4$}
		\label{fig:sequence_length_over_full_time:4}
	\end{subfigure} \\
	\begin{subfigure}[b]{0.49\textwidth}
		\includegraphics[width=\textwidth]{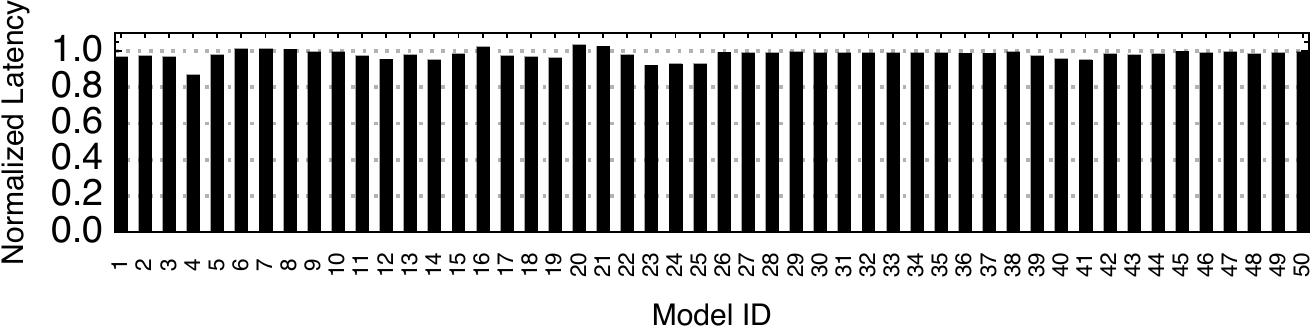}
		\caption{Benchmark Granularity=$5$}
		\label{fig:sequence_length_over_full_time:5}
	\end{subfigure}%
	\hfill
	\begin{subfigure}[b]{0.49\textwidth}
		\includegraphics[width=\textwidth]{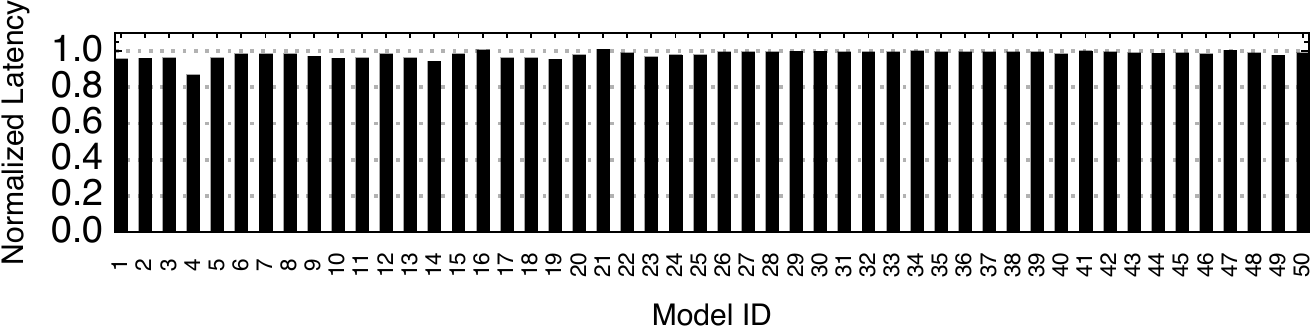}
		\caption{Benchmark Granularity=$6$}
		\label{fig:sequence_length_over_full_time:6}
	\end{subfigure}%
	\caption{The constructed model latency normalized to the model's end-to-end latency for the $50$ model in Table~\ref{tab:models_full} on \texttt{c5.2xlarge}.
		The benchmark granularity varies from $1$ to $6$. Sequence $1$ means each benchmark has one layer (layer granularity).}
	\label{fig:sequence_length_over_full_time}
\end{figure*}

\subsection{DL Model Performance Construction}\label{sec:perfcon}

\proj uses the performance of the layer sequences to construct an estimate to the end-to-end performance of the input model $M$.
To construct a performance estimate, the input model is parsed and goes through the same process \circledwhite{1} in Figure~\ref{fig:design}.
This creates a set of layer sequences.
The performance of each layer sequence is queried from the benchmark results ($P_{S_i}$ \ldots, $P_{S_k}$).
\proj supports both sequential and parallel performance construction.
Sequential performance construction is performed by summing up all of the resulting queried results, whereas parallel performance construction sums up the results along the critical path of the model.
Since current frameworks exhibit a sequential execution strategy (from Section~\ref{sec:perf}), sequential performance construction is used within \proj by default.
Other performance construction can be easily added to \proj to accommodate different framework execution strategies. %

\section{Evaluation}\label{sec:eval}

As the benchmark generation is automated and is based on a set of well-defined and consistent rules, \proj{} reduces the effort of developing and maintaining DL benchmarks.
Thus relieving the pressure of selecting representative DL models.
This section, therefore, focuses on demonstrating \proj{} is valid in terms of performance construction accuracy and benchmarking time speedup.
We explore the effect of benchmark granularity on the constructed performance estimation as well as the benchmarking time.
We evaluated \proj{} with $50$ DL models (listed in Table~\ref{tab:models_full}) using MXNet (v$1.5.1$ using MKL v$2019.3$) on $4$ different Amazon EC2 instances.
These systems are recommended~\cite{awscpu} by Amazon for DL inference and are listed in Table~\ref{tab:systems}.
Each model or benchmark is run $100$ times and the $20\%$ trimmed mean is reported.

\begin{table}
	\resizebox{0.45\textwidth}{!}{
		\begin{tabular}{ l l l r r }
			\toprule
			\centering \textbf{Instance} & \textbf{CPUS}           & \textbf{Memory (GiB)} & \textbf{\$/hr} \\ \midrule
			\texttt{c5.xlarge}           & 4 Intel Platinum 8124M  & 8GB                   & 0.17           \\
			\texttt{c5.2xlarge}          & 8 Intel Platinum 8124M  & 16GB                  & 0.34           \\
			\texttt{c4.xlarge}           & 4 Intel Xeon E5-2666 v3 & 7.5GB                 & 0.199          \\
			\texttt{c4.2xlarge}          & 8 Intel Xeon E5-2666 v3 & 15GB                  & 0.398          \\
			\bottomrule
		\end{tabular}%
	}
	\caption{ Evaluations are performed on the $4$ Amazon EC2 systems listed.
		The \texttt{c5.*} operate at 3.0 GHz, while the \texttt{c4.*} systems operate at 2.9 GHz.
		The systems are ones recommended by Amazon for DL inference.}
	\label{tab:systems}
	\vspace{-10pt}
\end{table}

\begin{figure}
	\centering
	\includegraphics[width=0.48\textwidth]{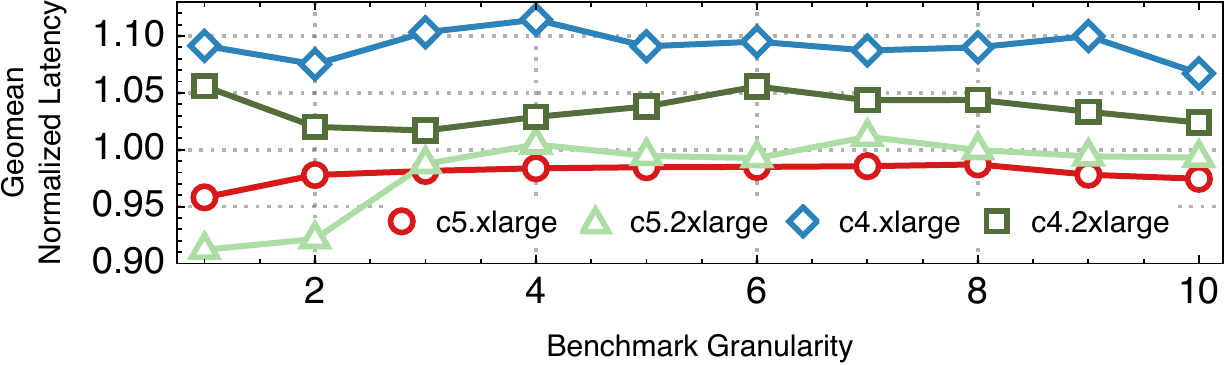}
	\caption{
		The geometric mean of the normalized latency (constructed vs end-to-end latency) of all the $50$ models on the $4$ systems with varying benchmark granularity from $1$ to $10$.
	}
	\label{fig:across_systems}
\end{figure}

\subsection{Performance Construction Accuracy}\label{sec:eval:perfcons}

We first ran the models on the $4$ systems to understand their performance characteristics, as shown in Figure~\ref{fig:end_to_end}.
Then using \proj, we constructed the latency estimate of the models based on the performance of their layer sequence benchmarks.
Figure \ref{fig:sequence_length_over_full_time} shows the constructed model latency normalized to the model's end-to-end latency for all the models with varying benchmark granularity from $1$ to $6$ on \texttt{c5.2xlarge}.
We see that the constructed latency is a tight estimate of the model's actual performance across models and benchmark granularities.
E.g., for benchmark granularity $G=1$, the normalized latency ranges between $82.9\%$ and $98.1\%$ with a geometric mean of $91.8\%$.

Figure~\ref{fig:across_systems} shows the geometric mean of the normalized latency (constructed vs end-to-end latency) of all the $50$ models across systems and benchmark granularities.
Overall, the estimated latency is within $5\%$  (e.g. $G = 3, 5, 9, 10$) to $11\%$ ($G = 1$) of the model end-to-end latency across systems.
This demonstrates that \proj{} provides a tight estimate to input models' actual performance.

\subsection{Benchmarking Time Speedup}\label{sec:eval:systems}

\begin{figure}[t]
	\centering
	\includegraphics[width=0.48\textwidth]{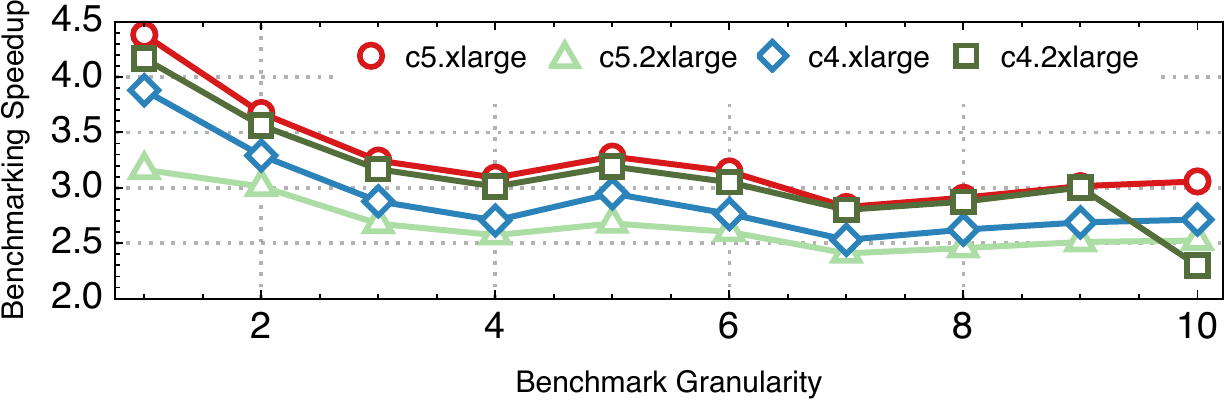}
	\caption{
		The speedup of total benchmarking time for the all the models across systems and benchmark granularities.
	}
	\label{fig:benchmark_speedup}
\end{figure}

\proj decreases the benchmarking time by only evaluating the unique layer sequences within and across models.
Recall from Section~\ref{sec:sharing} that for all the $50$ models, the total number of layers is $10815$, but only $1529$ are unique (i.e. $14\%$ are unique).
Figure~\ref{fig:benchmark_speedup} shows the speedup of the total benchmarking time across systems as benchmark granularity varies.
The benchmarking time speedup is calculated as the sum of the end-to-end latency of all models divided by the sum of the latency of all the generated benchmarks.
Up to $4.4\times$ benchmarking time speedup is observed for $G = 1$ on the \texttt{c5.xlarge} system.
The speedup decreases as the benchmark granularity increases.
This is because as the benchmark granularity increases, the chance of having repeated layer sequences within and across models decreases.

Figure~\ref{fig:sequence_length_over_full_time} and Figure~\ref{fig:benchmark_speedup} suggest a trade-off exists between the performance construction accuracy and benchmarking time speedup and the trade-off is system-dependent.
For example, while $G=1$ (layer granularity model decomposition and construction) produces the maximum benchmarking time speedup, the constructed latency is slightly less accurate comparing to other $G$ values on the systems.
Since this accuracy loss is small, overall, $G=1$ is a good choice of benchmark granularity configuration for \proj{} given the current DL software stack on CPUs.

\section{Related Work}\label{sec:related}

\paragraph{DL Benchmarking:}
To characterize the performance of DL models, both industry and academia have invested heavily in developing benchmark suites that characterize models and systems.
The benchmarking methods are either end-to-end benchmarks (performing user-observable latency measurement on a set of representative DL models~\cite{mlperf, dawnbench, aimatrix}) or are micro-benchmarks~\cite{convbench,deepbench, aimatrix} (isolating common kernels or layers that are found in models of interest).
The end-to-end benchmarks target end-users and measure the latency or throughput of a model under a specific workload scenario.
The micro-benchmark approach, on the other hand, distills models to their basic atomic operations (such as dense matrix multiplies, convolutions, or communication routines) and measures their performance to guide hardware or software design improvements~\cite{dong2018characterizing,Merck:2019:CED:3332186.3333049}.
While both approaches are valid and have their use cases, their benchmarks are manually selected and developed.
Thus, curation and maintaining these benchmarks facing requires significant effort and, in the case of lack of maintenance, becomes less representative of real-world models.

\proj{} complements the DL benchmarking landscape as it introduces a novel benchmarking methodology which reduces the effort of developing, maintaining, and running DL benchmarks.
\proj{} relieves the pressure of selecting representative DL models and copes well with the fast-evolving pace of DL models.
\proj{} automatically decomposes DL models into runnable networks and generates micro-benchmark based on these networks.
Users can specify the benchmark granularity.
At the two extremes, when the granularity is $1$ a layer-based micro-benchmark is generated, whereas when the granularity is equal to the number of layers within the model then an end-to-end network is generated.
To the best of our knowledge, there has been no previous work solving the same problem and we are the first to propose such as design.

\paragraph{Others:}
Previous work~\cite{elsken2019neural,dai2019chamnet} has also decomposed DL models into layers to guide performance-aware  neural architecture search.
\proj{} focuses on model performance and aims to reduce benchmarking effort.
\proj{} shares a similar spirit as synthetic benchmark generation.
There is work about synthetic benchmark generation in other domains~\cite{hutton2002automatic}, however, to the authors' knowledge, there has been no work on applying or specializing the synthetic benchmark generation to the DL domain.

\section{Discussion and Future Work}\label{sec:future}

\paragraph{Generating Overlapping Benchmarks}---
The current design only considers non-overlapping layer sequences during benchmark generation.
This may inhibit some types of optimizations (such as layer fusion).
A solution requires a small tweak to Algorithm~\ref{alg:model_subsequence} where we increment the \textit{begin} by $1$ rather than the end index of the \texttt{SplitModel} algorithm (line $7$).
A small modification is also needed within the performance construction step to pick the layer sequence resulting in the smallest latency.
Future work would explore the design space when generated benchmarks can overlap.

\paragraph{Adapting to Framework Evolution}---
The current \proj{} design is based on the observation that current DL frameworks do not execute data-independent layers in parallel.
Although \proj{} supports both sequential and parallel execution (Section~\ref{sec:perfcon}), as DL frameworks start to have some support of parallel execution of data-independent layers, the current design may be no longer suitable and needs to be adjusted.
To adapt \proj{} to this evolution of frameworks, one can adjust \proj to take user-specified parallel execution rules.
\proj{} can then use the parallel execution rules to make a more accurate model performance estimation.

\paragraph{Future Work}---
While this work focuses on CPUs, we expect the design to hold for GPUs as well.
Future work would explore the design for GPUs.
We are also interested in other use cases that are afforded by the \proj{} design --- model/system comparison and advising for the cloud~\cite{rausch2019towards,baylor2017tfx,hummer2019modelops}.
For example, it is common to ask questions such as, \textit{given a DL model which system should I use? or given a system and a task, which model should I use?}
These questions might come from a user who wants to pick the system that's ideal for the target model, a cloud provider who needs to schedule or migrate DL tasks, etc.
Using \proj, the system provider can curate a continuously updated database of the generated benchmarks results across its system offerings.
The system provider can then perform a performance estimate of the user's DL model (without running it) and give suggestions as to which system to choose.

\section{Conclusion}\label{sec:conc}

The fast-evolving landscape of DL posts considerable challenges in the DL benchmarking practice.
While benchmark suites are under pressure and are struggling to be agile, up-to-date, and representative, we take a different approach and propose a novel benchmarking design aimed at relieving this pressure.
Leveraging the key observations that layers are the performance building block of DL models and the layer repeatability within and across models,
\proj{} automatically generates composable benchmarks that reduce the effort of developing, maintaining, and running DL benchmarks.
Through the evaluation of state-of-the-art models on representative systems, we demonstrated that \proj{} copes with the fast-evolving pace of DL models.

% \clearpage
% \newpage

% \bibliographystyle{IEEEtran}

% \bibliographystyle{acm}
\bibliographystyle{ACM-Reference-Format}
\bibliography{main}

% \clearpage
% \newpage

% \input{sections/99-supplementary.tex}

\end{document}